\documentclass{article}

\usepackage{microtype}
\usepackage{graphicx}
\usepackage{subfigure}
\usepackage{booktabs} 
\usepackage{amsthm}
\theoremstyle{definition}
\newtheorem{definition}{Definition}[section]
\newtheorem{theorem}{Theorem}[section]

\newtheorem{lemma}[theorem]{Lemma}
\usepackage{amssymb}
\usepackage{float}
\usepackage{amsmath}
\usepackage{multirow}
\newcommand{\tabincell}[2]{\begin{tabular}{@{}#1@{}}#2\end{tabular}}

\usepackage{hyperref}

\usepackage[accepted]{icml2021}

\icmltitlerunning{Boosting Certified $\ell_\infty$ Robustness with EMA Method and Ensemble Model}

\begin{document}

\twocolumn[
\icmltitle{Boosting Certified $\ell_\infty$ Robustness\\
with EMA Method and Ensemble Model}



\icmlsetsymbol{equal}{*}

\begin{icmlauthorlist}
\icmlauthor{Binghui Li}{equal,to}
\icmlauthor{Shiji Xin}{equal,to}
\icmlauthor{Qizhe Zhang}{equal,to}
\end{icmlauthorlist}

\icmlaffiliation{to}{School of EECS, Peking University}

\icmlcorrespondingauthor{Qizhe, Zhang}{theia@pku.edu.cn}

\icmlkeywords{certified robustness, adversarial robustness, $\ell_\infty$-dist neuron, EMA method, model ensemble}

\vskip 0.3in
]



\printAffiliationsAndNotice{\icmlEqualContribution}

\begin{abstract}
The neural network with $1$-Lipschitz property based on $\ell_\infty$-dist neuron has a  theoretical guarantee in certified $\ell_\infty$ robustness. However, due to the inherent difficulties in the training of the network, the certified accuracy of previous work is limited. In this paper, we propose two approaches to deal with these difficuties. Aiming at the characteristics of the training process based on $\ell_\infty$-norm neural network, we introduce the EMA method to improve the training process. Considering the randomness of the training algorithm, we propose an ensemble method based on trained base models that have the $1$-Lipschitz property and gain significant improvement in the small parameter network. Moreover, we give the theoretical analysis of the ensemble method based on the $1$-Lipschitz property on the certified robustness, which ensures the effectiveness and stability of the algorithm.
Our code is available at \url{https://github.com/Theia-4869/EMA-and-Ensemble-Lip-Networks}.
\end{abstract}

\section{Introduction}
\label{introduction}
`Certified robustness is a strong guarantee of the robustness of neural networks. Previous studies have used techniques like adversarial training, interval bound relaxation and randomized smoothing. The work \cite{lipnet} by Zhang et al., however, proposed a simple yet effective way to deal with certified robustness by using $\ell_\infty$-dist neurons, which provides an efficient framework to solve $\ell_\infty$-norm restricted perturbation and gives the robust generalization bound based on the $1$-Lipschitz property of the network. The $1$-Lipschitz property implies that achieving good robustness accuracy on the training dataset is the key to achieving good practical results for this framework. The proposed network structure has a theoretical guarantee and reached the state of the art on standard test sets, but its performance is still limited and the structures are rather simple. We believe this is due to the sparsity of its gradient which makes the network hard to train. In this report, we focused on improving the expressing ability of networks built upon $\ell_\infty$-dist neurons while preserving its robustness. We present our results on improving the certified robustness of neural networks along with a few unimplemented ideas. In addition, we provide formal formulations for these methods.

First, we try to construct advanced network structures using $\ell_\infty$-dist neurons. As the accuracy rate is an upper bound of the robust/certified accuracy rate, we believe that if a more complex network structure could improve the standard accuracy rate, the robust accuracy and certified accuracy are likely to be improved. However, the results have shown that the realization of the complex network structure (LeNet/AlexNet/VGGNet) based on $\ell_\infty$-dist neurons did not meet our expectations. We analyze the possible reasons. Then we try to train the network through non-gradient-based optimization algorithms. After analyzing the key points of the training process in detail, we use the improved optimization algorithm (EMA) to train the network. Finally, inspired by ensemble learning \cite{more}, we implement the LeNet-based ensemble model and compare it with the pure LeNet structure. The result has shown a significant improvement, with a certified robustness accuracy rate close to the state-of-the-art level. 

Our contributions are summarized as follows:
\begin{itemize}
    \item We prove that the training of $\ell_\infty$-dist networks through first-order optimization methods is intrinsically hard, which coincides with the results from our attempt at building advanced network structures.
    \item We use exponential moving average (EMA) to train the network and get better performance.
    \item We propose ensemble methods and tested them on multiple network structures. The ensemble methods improve small networks significantly.
    \item We provide a theoretical analysis of ensemble methods, which gives a guarantee for the effectiveness and stability of the method.
\end{itemize}

\section{Methods}

\subsection{Advanced Network Structures}
As the network structures presented in the original paper are simple and the standard test accuracy is pretty low, we try to improve its expressive power by constructing several advanced network structures using $\ell_\infty$-dist neurons including LeNet \cite{lenet}, AlexNet \cite{alexnet}, and VGGNet \cite{vgg}. The original form of these networks has reached the state of the art on clean test sets.

\subsection{Training via Non-gradient Based Optimizer}
Training via gradient descent incurs well-known problems like vanishing and exploding gradients, where the former is an inherent weakness of $\ell_\infty$-dist nets as it has a sparse gradient. The MAdam optimizer \cite{madam} attempts to alleviate this problem by using multiplicative weight update to train the network, which proved to be efficient on deep neural networks. We use MAdam to train some of the networks we built.

\subsection{Exponential Moving Average}
Exponential Moving Average is an effective technique in training that performs an exponential average of the weights traversed by a stochastic optimizer with a modified learning rate schedule. We implement this technique on our models.

Formally, an exponential moving average $\theta^{\prime}$ of the model parameters $\boldsymbol{\theta}$ with a decay rate $\tau$ (i.e., $\boldsymbol{\theta}^{\prime} \leftarrow \tau \cdot \boldsymbol{\theta}^{\prime}+(1-\tau) \cdot \boldsymbol{\theta}$ at each
training step). During evaluation, the weighted parameters $\boldsymbol{\theta}^{\prime}$ are used instead of the trained parameters $\boldsymbol{\theta}$. 

\subsection{Model Ensemble}
Recall the nature of adversarial robustness, when the training is completed, samples in the instance space are divided into different categories by the classifier curve (the image of the neural network function in the high-dimensional space). Although the theoretical analysis of the dataset shows there is a sufficiently large gap between data points of different categories ($188$ for MNIST, $81$ for Fashion-MNIST, $54$ for CIFAR-10), for the classifier model based on deep neural networks, this gap is fragile. In other words, we can perturb the boundary of the classification curve with a small $\ell_\infty$-norm-limited perturbation to make it misclassify.

Thus, we propose a theoretically guaranteed model based on the ensemble. Intuitively, if we use multiple training models to implement the ensemble model, we could maintain the overall prediction accuracy at a certain level (In fact, better results are usually achieved, but it is out of our concern), and the $1$-Lipschitz property of the overall model function will also be guaranteed. 

What's more, because our pre-trained model is heterogeneous in a certain sense, under the constraint of $\ell_\infty$ norm, small disturbances will no longer cause the instance to be misclassified easily. Formally, for an example $\mathbf{x}$, for each model $\mathbf{M}_i$ ($1\leqslant i \leqslant m$) participating in the ensemble, there is an optimal $\ell_\infty$ restricted disturbance $\delta_i$ to make $\mathbf{x}+\delta_i$ misclassified as much as possible, but due to the heterogeneity of the model, the classification curve of each model is at the boundary with different refinement structures, these optimal disturbance vectors will also be different. Thus, it would be hard for a universal perturbation $\delta$ to make most models participating in ensemble misclassify $\mathbf{x}+\delta$, which leads to better robustness.

\section{Theoretical Analysis}

\subsection{Defects in the Training Process of Neural Network Based on $\ell_\infty$-dist Neurons and Advantages of EMA Method}

\begin{theorem}
\label{thm: 1}
    If $f_n(\mathbf{x}) \rightrightarrows \|\mathbf{x}\|_\infty$, we have 
    \begin{equation}
        \nabla f_n(\mathbf{x}) \rightrightarrows \nabla \|\mathbf{x}\|_\infty
    \end{equation}
\end{theorem}
The proof of this theorem is straightforward.

This simple theorem shows that no matter what kind of function family we use to approximate $\ell_\infty$ function ($\ell_p$ used in \cite{lipnet}), when the approximation sequence is very close to $\ell_\infty$, the first-order  optimization method will face the dilemma of sparse gradient. In fact, the training algorithm is reduced to the coordinate descent method.

This result inspires us to use a training algorithm with early advantages. Here we use the EMA algorithm. EMA smoothes the training process and could greatly improve the performance and robustness of the model in the early stage of training, but may lead to training stagnation in the later stage due to slow updating. Fortunately, this characteristic is in line with our training strategy.

\subsection{Bounding Certified Test Error of the Ensemble Model}

Now, we analyze the generalization of ensemble-based classifiers with $1$-Lipschitz guarantee from a theoretical perspective. We will give a generalization bound for the robust test error of $\ell_{\infty}$-dist nets. Let $(\mathbf{x}, y)$ be an instance-label pair where $\mathbf{x}\in\mathbb{K}$ and $y\in\{1,-1\}$ and denote $\mathcal{D}$ as the distribution of $(\mathbf{x}, y)$. We define ensemble classification function $G(\mathbf{x})=\sum\limits_{i=1}^{m}w_i g_i(\mathbf{x})$, where $w_i$ is weight, $g_i(\mathbf{x}): \mathbb{R}^{d_{\text {input }}} \rightarrow \mathbb{R}$ is the base function, we use $\operatorname{sign}(G(\mathbf{x}))$ to denote the ensemble classifier. 

~\\

\begin{definition}
The $r$-robust test error $\gamma_{r}$ of a classifier $G$ is defined as
\begin{equation}
    \gamma_{r}=\mathbb{E}_{(\mathbf{x}, y) \sim \mathcal{D}}\left[\sup _{\left\|\mathbf{x}^{\prime}-\mathbf{x}\right\|_{\infty} \leqslant r} \mathbb{I}\left[y G\left(\mathbf{x}^{\prime}\right) \leqslant 0\right]\right]
\end{equation}
\end{definition}

\begin{theorem}
\label{thm: 2}
Let $\mathbb{F}$ denote the set of all $g$ represented by an $\ell_{\infty}$-dist net with width at most $W$ and depth at most $L$. Let $\mathbb{H}$ denote the set of all $G$ combined by base model $g \in \mathbb{F} $. For every $t>0$, with probability at least $1-2 e^{-2 t^{2}}$ over the random drawing of $n$ samples, for all $r>0$ and $G \in \mathbb{H}$ we have that
\begin{align}
\gamma_{r} \leqslant \inf _{\delta \in(0,1]}\left[\frac{1}{n} \sum\limits_{i=1}^{n} \mathbb{I}\left[y_{i} G\left(\mathbf{x}_{i}\right) \leqslant \delta+r\right]+\tilde{O}\left(\frac{L W^{2}}{\delta \sqrt{n}}\right)\right. \\
\left.+\left(\frac{\log \log _{2}\left(\frac{2}{\delta}\right)}{n}\right)^{\frac{1}{2}}\right]+\frac{t}{\sqrt{n}} \notag
\end{align}
\end{theorem}

On the right side of the generalization boundary inequality of the above theorem, the first term is training robust error, and the second term reflects the size of the base network used. It can be interpreted that the use of multiple base classification models does not increase the generalization bound on the network size item, but the ensemble method makes the model have lower training robust error, so our model has a better theoretical guarantee.

\subsection{Bounding Certified Training Error of the Ensemble Model}

Now, we turn our attention to the robust training error analysis of the ensemble model. Firstly, we formally define the certified training error of the ensemble model, and analyze its relationship with the number of base models $m$.

\begin{definition}
The $r$-certified test error $\hat{\gamma_{r}}$ of a classifier $G$ is defined as
\begin{equation}
    \hat{\gamma_{r}}=\frac{1}{n} \sum\limits_{i=1}^{n} \left[\sup _{\left\|{\mathbf{x}_i}^{\prime}-\mathbf{x}_i\right\|_{\infty} \leqslant r} \mathbb{I}\left[y_i G\left({\mathbf{x}_i}^{\prime}\right) \leqslant 0\right]\right]
\end{equation}
\end{definition}

Given dataset $S=\{(\mathbf{x}_1,y_1),(\mathbf{x}_2,y_2),...,(\mathbf{x}_n,y_n)\}$, due to the randomness of the training algorithm, we can assume that the trained model $g$ satisfies a distribution $D(S)$, which all make $y_i(g_j(\mathbf{x}_i))>0$. And we draw $g_1,g_2,\ldots,g_m$ from $D(S)$ to construct the ensemble model $G$. Then we assume the sample margin $\rho_{i,j}=y_i(g_j(\mathbf{x}_i))\sim \tilde{\rho_{i}}$, $\mathbb{E}\tilde{\rho_i} = \mu_i$. Because the model is homogeneous, we can consider that $\rho_i \in [0,1]$.

\begin{theorem}
\label{thm: 3}
With probability at least $1-t$ over the random drawing of $g_1,g_2,\ldots,g_m$ samples, we have
\begin{equation}
    \hat{\gamma_{r}}\leqslant \frac{1}{n} \sum\limits_{i=1}^{n}\mathbb{I}\left[r\geqslant \mu_i-\sqrt{\frac{\log(n/t)}{2m}}\right]
\end{equation}
\end{theorem}

The above theorem shows that when we increase the number of base classifiers $m$ participating in the ensemble, the certified training error will have a better theoretical upper bound, which is consistent with our experimental results. Combined with Theorem \ref{thm: 2}, we give a theoretical upper bound of $m$-related certified test error based on the ensemble model, which is obtained by us first.

\section{Experiments \& Results}
We evaluate our methods by measuring classification accuracy (standard, robust, and certified) on a number of common image classification benchmarks including MNIST, Fashion-MNIST, CIFAR-10, and CIFAR-100. For the network, we experiment with five commonly used architectures constructed by $\ell_\infty$-dist neurons: MLP, Conv, LeNet, AlexNet, and VGGNet. We use model ensemble to improve performance and robustness. We also use the training trick exponential moving average to improve robustness.
\subsection{Classification Accuracy}
We use “Standard”, “Robust” and “Certified” as abbreviations of standard (clean) test accuracy, robust test accuracy under PGD attack, and certified test accuracy. All the numbers are reported in percentage. We use “PARAMs” to denote the total number of parameters in model. Our training configurations are the same as configurations in \cite{lipnet}. Classification accuracy results are listed in TABLE \ref{table: 1}. Note that our methods outperform $\ell_\infty$-dist Net on both MNIST and CIFAR-10 datasets, and the results are also comparable to that $\ell_\infty$-dist Net on Fashion-MNIST dataset.

\begin{table*}[ht]
\centering
\begin{tabular}{c|cc|ccc}
\hline
Dataset                                        & Method                                     & PARAMs    & Test           & Robust         & Certified      \\ \hline
\multirow{5}{*}{\tabincell{c}{MNIST\\($\epsilon$=0.3)}}         & $\ell_\infty$-dist Net                     & 82708490  & 98.54          & 93.10          & 92.61          \\
                                               & $\ell_\infty$-dist Net+MLP                 & 83970826  & 98.60          & 93.62          & 92.92          \\
                                                                                              & $\ell_\infty$-dist Net+MLP+EMA             & 83970826  & 98.64          & 93.88          & $\mathbf{93.14}$ \\
                                               & ensemble $\ell_\infty$-dist Net*5+EMA     & 413542450 & $\mathbf{98.92}$ & $\mathbf{94.14}$ & 92.42          \\
                                               & ensemble $\ell_\infty$-dist Net+MLP*5+EMA & 419854130 & 98.64          & 94.99          & 92.77          \\ \hline
\multirow{5}{*}{\tabincell{c}{Fashion-MNIST\\($\epsilon$=0.1)}} & $\ell_\infty$-dist Net                     & 82708490  & 87.91          & 79.62          & 77.48          \\
                                               & $\ell_\infty$-dist Net+MLP                 & 85284362  & 87.91          & $\mathbf{80.89}$ & $\mathbf{79.23}$ \\
                                               & $\ell_\infty$-dist Net+MLP+EMA             & 85284362  & 88.42          & 80.47          & 79.01          \\
                                               & ensemble
                                               $\ell_\infty$-dist Net*5+EMA     & 413542450 & $\mathbf{88.86}$ & 80.14          & 77.16          \\
                                               & ensemble $\ell_\infty$-dist Net+MLP*5+EMA & 426421810 & 88.76          & 80.86          & 78.47          \\ \hline
\multirow{3}{*}{\tabincell{c}{CIFAR-10\\($\epsilon$=8/255)}}    & $\ell_\infty$-dist Net                     & 120637450 & $\mathbf{56.77}$ & 39.58          & 32.72          \\
                                               & $\ell_\infty$-dist Net+MLP                 & 123213322 & 51.04          & 38.22          & 35.17          \\
                                               & $\ell_\infty$-dist Net+MLP+EMA             & 123213322 & 52.10          & $\mathbf{40.43}$ & $\mathbf{35.42}$ \\ \hline
\end{tabular}
\caption{Comparison of our results with $\ell_\infty$-dist Net.}
\label{table: 1}
\end{table*}

\subsection{Advanced Network Structures}
We implement $1$-Lipschitz versions of three commonly used CNN architectures with $\ell_\infty$-dist neurons: LeNet, AlexNet, and VGGNet. We compare them to $\ell_\infty$-dist Net and ConvNet on CIFAR-10. The expressing ability of advanced network structure is improved, but because the gradient is difficult to propagate under $\ell_\infty$-norm, the parameters of the deep network are difficult to be updated, which leads to the decline of certified and robust accuracy of the network. But using $\ell_\infty$-dist neurons to build advanced CNN is a good attempt, and the lightweight network $\ell_\infty$-dist LeNet with very few parameters still achieves a good result, which is even comparable to that of $\ell_\infty$-dist Net and ConvNet whose parameters are 2-3 orders of magnitude higher. Results are listed in TABLE \ref{table: 2}.

\begin{table*}[ht]
\centering
\begin{tabular}{c|cc|ccc}
\hline
Dataset                    & Method                         & PARAMs    & Test  & Robust & Certified \\ \hline
\multirow{10}{*}{\tabincell{c}{CIFAR-10\\($\epsilon$=8/255)}} & $\ell_\infty$-dist Net         & 120637450 & 56.77 & 39.58  & 32.72     \\
                           & $\ell_\infty$-dist Net+MLP     & 123213322 & 51.04 & 38.22  & 35.17     \\
                           & $\ell_\infty$-dist ConvNet     & 51256874  & 54.12 & 33.98  & 29.47     \\
                           & $\ell_\infty$-dist ConvNet+MLP & 51776554  & 48.90 & 36.13  & 33.72     \\
                           & $\ell_\infty$-dist LeNet       & 117420    & 46.71 & 30.92  & 25.21     \\
                           & $\ell_\infty$-dist LeNet+MLP   & 703660    & 40.83 & 31.64  & 29.54     \\
                           & $\ell_\infty$-dist AlexNet     & 6613802   & 47.36 & 28.65  & 23.42     \\
                           & $\ell_\infty$-dist AlexNet+MLP & 7662890   & 40.40 & 31.51  & 29.50     \\
                           & $\ell_\infty$-dist VGGNet         & 26172106  & 54.46 & 34.90  & 24.46     \\
                           & $\ell_\infty$-dist VGGNet+MLP     & 26827978  & 40.02 & 30.86  & 28.56     \\ \hline
\end{tabular}
\caption{Comparison of our $\ell_\infty$-dist CNN with $\ell_\infty$-dist MLP and Conv.}
\label{table: 2}
\end{table*}

\subsection{Non-gradient Based Optimizer}
We use non-gradient based optimizer MAdam to train two $\ell_\infty$-dist CNN architectures: LeNet and LeNet+MLP on CIFAR-10 and compare them with the same architectures trained with AdamW optimizer, which is also used in \cite{lipnet}. In practice, AdamW optimizer outperforms MAdam in both models. We list the results in TABLE \ref{table: 3}.

\begin{table*}[ht]
\centering
\begin{tabular}{c|cc|ccc}
\hline
Dataset                   & Method                              & PARAMs & Test  & Robust & Certified \\ \hline
\multirow{4}{*}{\tabincell{c}{CIFAR-10\\($\epsilon$=8/255)}} & $\ell_\infty$-dist LeNet(AdamW)     & 117420 & $\mathbf{44.31}$ & $\mathbf{28.65}$  & $\mathbf{24.21}$     \\
                          & $\ell_\infty$-dist LeNet(MAdam)     & 117420 & 37.11 & 23.89  & 17.61     \\ \cline{2-6} 
                          & $\ell_\infty$-dist LeNet+MLP(AdamW) & 703660 & $\mathbf{33.15}$ & $\mathbf{27.54}$  & $\mathbf{25.36}$     \\
                          & $\ell_\infty$-dist LeNet+MLP(MAdam) & 703660 & 30.82 & 25.07  & 23.67     \\ \hline
\end{tabular}
\caption{Comparison of AdamW and MAdam optimizer.}
\label{table: 3}
\end{table*}

\subsection{Model Ensemble}
We use model ensemble to train 10 $\ell_\infty$-dist LeNet models on dataset MNIST, Fashion-MNIST, CIFAR-10, and CIFAR-100, and compare them with the performance of single model with same architecture on the same dataset. Note that ensemble models are superior to single models in almost all accuracy tests and show significant improvement in certified accuracy. We list the results in TABLE \ref{table: 4}.

\begin{table*}[ht]
\centering
\begin{tabular}{c|cc|ccc}
\hline
Dataset & Method                                    & PARAMs  & Test  & Robust & Certified \\ \hline
                                                       & $\ell_\infty$-dist LeNet+MLP              & 703360  & 88.08 & 75.33  & 75.82     \\
\multirow{-2}{*}{MNIST($\epsilon$=0.3)}                & ensemble $\ell_\infty$-dist LeNet+MLP*10 & 7033600 & $\mathbf{92.04}$ & $\mathbf{80.21}$  & $\mathbf{78.75}$     \\ \hline
                                                       & $\ell_\infty$-dist LeNet+MLP              & 703360  & 76.86 & 69.47  & 67.39     \\
\multirow{-2}{*}{Fashion-MNIST($\epsilon$=0.1)}        & ensemble $\ell_\infty$-dist LeNet+MLP*10 & 7033600 & $\mathbf{82.84}$ & $\mathbf{75.46}$  & $\mathbf{73.43}$     \\ \hline
                                                       & $\ell_\infty$-dist LeNet+MLP              & 703660  & 40.83 & $\mathbf{31.64}$  & 29.54     \\
\multirow{-2}{*}{CIFAR-10($\epsilon$=8/255)}           & ensemble $\ell_\infty$-dist LeNet+MLP*10 & 7036600 & $\mathbf{40.89}$ & 31.38  & $\mathbf{33.42}$     \\ \hline
                                                       & $\ell_\infty$-dist LeNet+MLP              & 749830  & 13.89 & 9.90   & 7.85      \\
\multirow{-2}{*}{CIFAR-100($\epsilon$=8/255)}          & ensemble $\ell_\infty$-dist LeNet+MLP*10 & 7498300 & $\mathbf{19.38}$ & $\mathbf{13.09}$  & $\mathbf{10.77}$     \\ \hline
\end{tabular}
\caption{Comparison of ensemble model and single model.}
\label{table: 4}
\end{table*}

\section{Unimplemented Ideas}

\subsection{Another Convergence Sequence to $\ell_\infty$-norm}
\begin{figure}
    \centering
    \includegraphics[width=\linewidth]{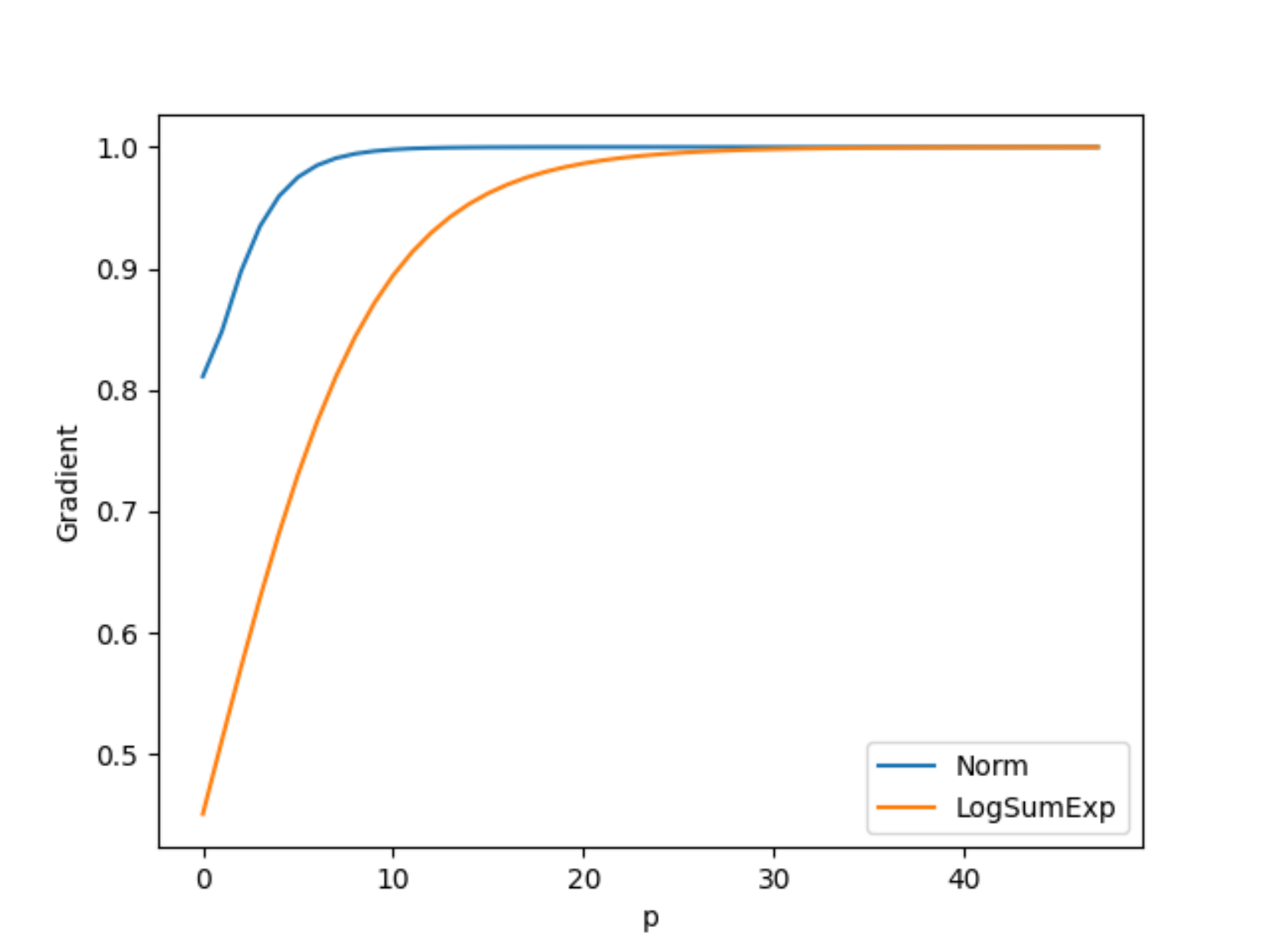}
    \caption{Gradients of $\ell_p$-norm and Log-Sum-Exp function}
    \label{fig:lse}
\end{figure}

In the original paper \cite{lipnet}, authors used $\ell_p$-dist neuron as a surrogate to the $\ell_\infty$-dist neuron to get a non-sparse approximation of the original gradient, which boosted the performance. Here we propose another approach by using the Log-Sum-Exp function:
\begin{equation}
    u(\mathbf{z}, \theta, p) = \frac{\log\sum_i\exp(p|z_i - w_i|)}{p} + b \quad (p>0)
\end{equation}
where $\theta = \{\mathbf{w}, b\}$ is the parameter set and $p$ is an adjustable parameter, the larger $p$ is, the closer the result is to the $\ell_\infty$ distance. Numerical experiments show this gives smoother gradient than $\ell_p$-norm which is better for training. Check Fig. \ref{fig:lse} for details.
\subsection{Residual Connection}
Residual connections offer a way to pass gradients directly, which alleviates the vanishing gradients problem and leads to good performance on very deep networks \cite{resnet}. As $\ell_\infty$-dist nets experience similar problems, we think residual connections may bring better performance. The test could be performed on $\ell_\infty$ version of  ResNet \cite{resnet} and WideResNet \cite{wideresnet}.
\begin{equation}
    u(\mathbf{z}, \theta) = c\mathbf{z} + (1-c)\|\mathbf{z}-\mathbf{w}\|_\infty + b, c\in [0, 1)
    \label{residual}
\end{equation}

\section{Conclusion}
In this paper, we analyze the intrinsic difficulties in the training process of neural networks based on $\ell_\infty$-norm neurons and propose two effective methods to enhance certified robustness. First, we use the EMA method to achieve the state of the art results. Then, based on the randomness of the training algorithm, we propose an ensemble method with a theoretical guarantee, which significantly improves the performance of a neural network that is much smaller than the one used in \cite{lipnet}. In the future, we believe the combination of a better ensemble method and models with more parameters and larger diversity could 
lead to better results.

\section*{Acknowledgment}

The authors would like to thank Prof. Liwei Wang and TAs for their dedicated work.


\bibliography{ref}
\bibliographystyle{icml2021}

\appendix
\section{Proof of Theorem \ref{thm: 2}}
Now we will prove Theorem \ref{thm: 2} by three lemmas. The first lemma shows the connection between the Rademacher complexity of the base hypothesis class and the ensemble (convex combination) model class. The second lemma shows that the ensemble model also has the $1$-Lipschitz property. Then we improve the third lemma by first and second lemmas to get the main theorem.
\begin{definition}[Rademacher Complexity]~\\
Given a sample $\mathbf{X}_{n}=$ $\left\{\mathbf{x}_{1}, \ldots, \mathbf{x}_{n}\right\} \in \mathbb{K}^{n}$, and a real-valued function class $\mathbb{F}$ on
$\mathbb{K}$, the Rademacher complexity of $\mathbb{F}$ is defined as
\begin{equation}
    R_{n}(\mathbb{F})=\mathbb{E}_{\mathbf{x}_{n}}\left(\frac{1}{n} \mathbb{E}_{\sigma}\left[\sup _{f \in \mathbb{F}} \sum_{i=1}^{n} \sigma_{i} f\left(\mathbf{x}_{i}\right)\right]\right)
\end{equation}
where $\sigma_{i}$ are drawn from the Rademacher distribution independently, i.e. $\mathbb{P}\left(\sigma_{i}=1\right)=\mathbb{P}\left(\sigma_{i}=-1\right)=\frac{1}{2}$. It's worth
noting that for any constant function $r$, $R_{n}(\mathbb{F})=R_{n}(\mathbb{F} \oplus r)$ where $\mathbb{F} \oplus r=\{f+r \mid f \in \mathbb{F}\}$.
\end{definition}

\begin{lemma}
\label{lemma:1}
Let $H$ be a set of functions mapping from $\mathcal{X}$ to $\mathbb{R}$. Then, for any sample $S$, we have
\begin{equation}
    \widehat{\Re}_{S}(\operatorname{conv}(H))=\widehat{\Re}_{S}(H)
\end{equation}
where $\widehat{\Re}_{S}(\cdot) $ denote the empirical Rademacher complexity. We can get the lemma directly using the linearity of expectation.
\end{lemma}

\begin{lemma}
\label{lemma:2}
Assume function $g \in \mathbb{F} $ is $1$-Lipschitz (with respect to $\ell_\infty$-norm), let $\mathbb{H}$ denote the set of all $G$ convex combined by base model $g \in \mathbb{F}  $.Then function $G \in \mathbb{H}$ is also $1$-Lipschitz (with respect to $\ell_\infty$-norm).

The lemma can be proved by the triangle inequality of $\ell_\infty$-norm and the $1$-Lipschitz property of the neural network based on $\ell_\infty$-norm.
\end{lemma}

\begin{lemma}[Theorem 4.2 in \cite{lipnet} by Zhang et al.]
\label{lemma:3}
Let $\mathbb{F}$ denote the set of all $g$ represented by an $\ell_{\infty}$-dist net with width at most $W$ and depth at most $L$. For every $t>0$, with probability at least $1-2 e^{-2 t^{2}}$ over the random drawing of $n$ samples, for all $r>0$ and $g \in \mathbb{F}$ we have that
\begin{align}
\gamma_{r} \leqslant \inf _{\delta \in(0,1]}\left[\frac{1}{n} \sum\limits_{i=1}^{n} \mathbb{I}\left[y_{i} g\left(\mathbf{x}_{i}\right) \leqslant \delta+r\right]+\tilde{O}\left(\frac{L W^{2}}{\delta \sqrt{n}}\right)\right. \\
\left.+\left(\frac{\log \log _{2}\left(\frac{2}{\delta}\right)}{n}\right)^{\frac{1}{2}}\right]+\frac{t}{\sqrt{n}} \notag
\end{align}
\end{lemma}

Finally, Theorem \ref{thm: 2} is a direct consequence by combing
Lemmas \ref{lemma:1}, \ref{lemma:2} and \ref{lemma:3}.

\section{Proof of Theorem \ref{thm: 3}}
In this section, we will prove Theorem \ref{thm: 3} using concentration inequality and union bound.

Firstly, we define the minimum $\ell_{\infty}$-norm perturbation $r_i$ w.r.t. $(\mathbf{x}_i,y_i)$ as
\begin{equation}
    r_i=\inf \left\{r>0 \left| \sup _{\left\|{\mathbf{x}_i}^{\prime}-\mathbf{x}_i\right\|_{\infty} \leqslant r} \mathbb{I}\left[y_i G\left({\mathbf{x}_i}^{\prime}\right) \leqslant 0\right]=1\right\}\right.
\end{equation}
Since $G$ has $1$-Lipschitz property, with the assumption in Theorem \ref{thm: 3}, we have
\begin{equation}
    r_i \geqslant \frac{1}{m} \sum\limits_{j=1}^{m} \rho_{i,j}
\end{equation}
As $\rho_i \in [0,1]$, we could apply Chernoff Bound to get
\begin{equation}
    \mathbb{P} \left(\mu_i-\frac{1}{m} \sum\limits_{j=1}^{m} \rho_{i,j} \geqslant \epsilon\right) \leqslant e^{-2 m \epsilon^2}
\end{equation}
By applying union bound, we have
\begin{equation}
    \mathbb{P} \left(\exists i \in [n], \mu_i-\frac{1}{m} \sum\limits_{j=1}^{m} \rho_{i,j} \geqslant \epsilon\right) \leqslant n e^{-2 m \epsilon^2}
\end{equation}
Thus, with probability at least $1-t$, we have
\begin{equation}
    \frac{1}{m} \sum\limits_{j=1}^{m} \rho_{i,j} \geqslant \mu_i-\sqrt{\frac{\log(n/t)}{2m}}, \forall i \in [n]
\end{equation}
Finally, from the definition of $\hat{\gamma_{r}}$, we have
\begin{align}
    \hat{\gamma_{r}}&=\frac{1}{n} \sum\limits_{i=1}^{n} \left[\sup _{\left\|{\mathbf{x}_i}^{\prime}-\mathbf{x}_i\right\|_{\infty} \leqslant r} \mathbb{I}\left[y_i G\left({\mathbf{x}_i}^{\prime}\right) \leqslant 0\right]\right]\\
    &=\frac{1}{n} \sum\limits_{i=1}^{n}\mathbb{I}\left[r \geqslant r_i\right] \\
    &\leqslant \frac{1}{n}\sum\limits_{i=1}^{n}\mathbb{I}\left[r \geqslant  \frac{1}{m}\sum\limits_{j=1}^{m} \rho_{i,j}\right] \\
    &\leqslant \frac{1}{n}\sum\limits_{i=1}^{n}\mathbb{I}\left[r\geqslant \mu_i-\sqrt{\frac{\log(n/t)}{2m}}\right]
\end{align}
With probability at least $1-t$.

\section{Ablation Study}
In this section, we conduct ablation experiments to view the
effects of different types of ensemble methods, different EMA factors and different data augmentations.
\subsection{Ensemble Method}
We try two different ensemble methods: Fusion and Voting. In Fusion, the output from all base estimators is aggregated as an average output; in Voting, the softmax normalization is conducted before taking the average of predictions from all base estimators. For both methods, we train 10 $\ell_\infty$-dist LeNet+MLP on CIFAR-10 dataset to see the performance. The results of comparison are presented in the TABLE \ref{table: 6}. We use Fusion ensemble in all previous experiments.

\begin{table}[ht]
\centering
\begin{tabular}{|c|c|c|c|}
\hline
Method & Test  & Robust & Certified \\ \hline
Fusion & 40.89 & 31.38  & 33.42     \\ \hline
Voting & 40.53 & 30.60  & 33.48     \\ \hline
\end{tabular}
\caption{Comparison of two different ensemble methods.}
\label{table: 5}
\end{table}

\subsection{EMA Factor}
Exponential Moving Average is an effective technique to improve performance, but the factor used for weight updating can have a big impact on the result. Here we try two different factors used in EMA: $0.99$ and $0.999$, both experiments are conducted on CIFAR-10 using $\ell_\infty$-dist MLP. The results of comparison are presented in the TABLE \ref{table: 6}. We finally choose the factor of $0.99$ in all previous experiments.

\begin{table}[ht]
\centering
\begin{tabular}{|c|c|c|c|}
\hline
EMA Factor & Test  & Robust & Certified \\ \hline
0.99       & 52.10 & 40.43  & 35.42     \\ \hline
0.999      & 51.70 & 38.80  & 35.20     \\ \hline
\end{tabular}
\caption{Comparison of EMA factor $0.99$ and $0.999$.}
\label{table: 6}
\end{table}

\subsection{Data Augmentation}
Data augmentation is critical for extracting high-dimensional features. In general, strong data augmentation can improve model robustness. Here we try three different data augmentations: WAUG (RandomCrop + RandomHorizontalFlip), MAUG (RandomCrop + RandomHorizontalFlip + RandomColorJitter + RandomGrayscale) and SAUG (RandomResizedCrop + RandomHorizontalFlip + RandomColorJitter + RandomGrayscale). All three experiments are conducted on CIFAR-10 using $\ell_\infty$-dist MLP and Conv, and the performances of all data augmentations are listed in TABLE \ref{table: 7}. We choose to use WAUG in all previous experiments.
\begin{table}[H]
\centering
\resizebox{9cm}{1.3cm}{
\begin{tabular}{|c|c|c|c|c|}
\hline
Method                                          & Data Aug & Test  & Robust & Certified \\ \hline
\multirow{3}{*}{$\ell_\infty$-dist Net+MLP}     & WAUG     & 51.04 & 38.22  & 35.17     \\ \cline{2-5} 
                                                & MAUG     & 45.59 & 36.13  & 33.41     \\ \cline{2-5} 
                                                & SAUG     & 41.94 & 35.29  & 32.11     \\ \hline
\multirow{3}{*}{$\ell_\infty$-dist ConvNet+MLP} & WAUG     & 48.90 & 36.13  & 33.72     \\ \cline{2-5} 
                                                & MAUG     & 46.80 & 35.61  & 33.15     \\ \cline{2-5} 
                                                & SAUG     & 40.88 & 33.20  & 30.62     \\ \hline
\end{tabular}
}
\caption{Comparison of different data augmentations.}
\label{table: 7}
\end{table}

\end{document}